\pdfoutput=1

\documentclass[11pt]{article}

\usepackage{acl}

\usepackage{times}
\usepackage{latexsym}

\usepackage[T1]{fontenc}

\usepackage[utf8]{inputenc}

\usepackage{microtype}

\usepackage{tikz}
\usetikzlibrary{positioning}
\usetikzlibrary{bayesnet}
\usetikzlibrary{arrows}
\usetikzlibrary{backgrounds}
\usepackage{color}
\usepackage{graphicx}
\usepackage{caption}
\usepackage{subcaption}

\usepackage{amsmath}
\usepackage{amssymb}

\usepackage{cleveref}

\usepackage{todonotes}
\usepackage{hhline}
\usepackage{colortbl}

\usepackage{tabularx}

\usepackage{relsize}
\usepackage{multirow}

\newcommand*\rot{\rotatebox{55}}

\title{Hierarchical Sketch Induction for Paraphrase Generation}

\author{{Tom Hosking \qquad Hao Tang \qquad Mirella Lapata} \\
  Institute for Language, Cognition and Computation \\
  School of Informatics, University of Edinburgh \\
  10 Crichton Street, Edinburgh EH8 9AB\\
  \texttt{tom.hosking@ed.ac.uk} \quad \texttt{hao.tang@ed.ac.uk} \quad \texttt{mlap@inf.ed.ac.uk}}

\begin{document}
\maketitle
\begin{abstract}
We propose a generative model of paraphrase generation, that encourages syntactic diversity by conditioning on an explicit syntactic sketch. We introduce Hierarchical Refinement Quantized Variational Autoencoders (HRQ-VAE), a method for learning decompositions of dense encodings as a sequence of discrete latent variables that make iterative refinements of increasing granularity. This hierarchy of codes is learned through end-to-end training, and represents fine-to-coarse grained information about the input. We use HRQ-VAE to encode the syntactic form of an input sentence as a path through the hierarchy, allowing us to more easily predict syntactic sketches at test time. Extensive experiments, including a human evaluation, confirm that HRQ-VAE learns a hierarchical representation of the input space, and generates paraphrases of higher quality than previous systems.
\end{abstract}

\section{Introduction}



Humans use natural language to convey information, mapping an abstract idea to a sentence with a specific surface form. A paraphrase is an alternative surface form of the same underlying semantic content. The ability to automatically identify and generate paraphrases is of significant interest, with applications in data augmentation \cite{iyyer-etal-2018-adversarial}, query rewriting, \cite{dong-etal-2017-learning-paraphrase} and duplicate question detection \cite{shah-etal-2018-adversarial}.

 While autoregressive models of language (including paraphrasing systems) predict one token at a time, there is evidence that in humans some degree of planning occurs at a higher level than individual words \cite{levelt_planning,martin2010planning}. Prior work on paraphrase generation has attempted to include this inductive bias by specifying an alternative surface form as additional model input, either in the form of target parse trees \cite{iyyer-etal-2018-adversarial,chen-etal-2019-controllable,sgcp2020}, exemplars \cite{meng2021conrpg}, or syntactic codes \cite{shu-etal-2019-generating, hosking-lapata-2021-factorising}. Most of these approaches suffer from an `all or nothing' problem: the target surface form must be fully specified during inference. However, predicting the complete syntactic structure is almost as difficult as predicting the sentence itself, negating the benefit of the additional planning step.

\begin{figure}[t]
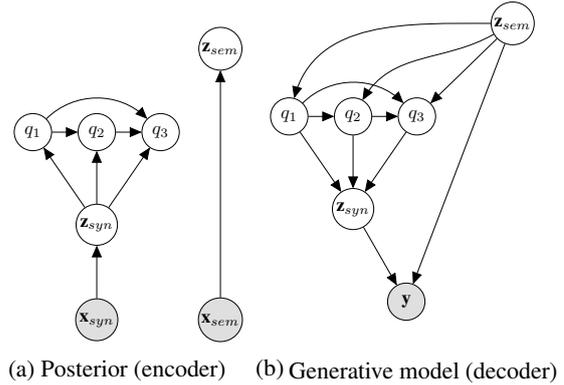

    \centering
    \begin{subfigure}{0.23\textwidth}
        \centering
        \scalebox{0.7}{
          \tikz{
         \node[obs] (xsyn) {$\textbf{x}_{syn}$};%
         \node[obs, right=of xsyn,xshift=0.5cm] (xsem) {$\textbf{x}_{sem}$};%
         \node[latent,above=of xsyn,xshift=0cm,yshift=0cm] (zsyn) {$\textbf{z}_{syn}$};
         \node[latent,above=of xsem,yshift=3.3cm] (zsem) {$\textbf{z}_{sem}$};
         \node[latent,above=of zsyn,xshift=-1.2cm] (q1) {$q_1$}; %
         \node[latent,above=of zsyn,xshift=-0cm] (q2) {$q_2$}; %
         \node[latent,above=of zsyn,xshift=1.2cm] (q3) {$q_3$}; %
        \edge{zsyn}{q1,q2,q3}   
        \edge{xsem}{zsem} 
        \edge{xsyn}{zsyn}  
        \edge{q1}{q2}
        \edge{q2}{q3}
        \draw[out=45,in=135, ->] (q1) edge (q3);
         }
         }
        \caption{Posterior (encoder)\hspace*{2ex}}
        \label{fig:gm_encoder}
    \end{subfigure}    \hspace*{-2ex}
    \begin{subfigure}{0.24\textwidth}
        \centering
        \scalebox{0.7}{
          \tikz{
         \node[obs] (y) {$\textbf{y}$};%
         \node[latent,above=of y,xshift=-1cm] (zsyn) {$\textbf{z}_{syn}$};
         \node[latent,above=of y,xshift=2cm,yshift=3.5cm] (zsem) {$\textbf{z}_{sem}$};
         \node[latent,above=of zsyn,xshift=-1.2cm] (q1) {$q_1$}; %
         \node[latent,above=of zsyn,xshift=-0cm] (q2) {$q_2$}; %
         \node[latent,above=of zsyn,xshift=1.2cm] (q3) {$q_3$}; 
        \edge{zsem}{q3}
        \edge{q1,q2,q3}{zsyn}    
        \edge{zsyn,zsem}{y}
         \edge{q1}{q2}
         \edge{q2}{q3}
        \draw[out=45,in=135, ->] (q1) edge (q3);
        \draw[out=180,in=75, ->] (zsem) edge (q1);
        \draw[out=210,in=60, ->] (zsem) edge (q2);
        
         }
         }\vspace*{.3cm}
        \caption{\mbox{\raisebox{2.2ex}[0pt]{\hspace*{3ex}Generative model (decoder)}}}
        \label{fig:gm_decoder}
    \end{subfigure}
    \caption{The generative models underlying our approach. Given some
      semantic content~$\textbf{z}_{sem}$, we predict a hierarchical
      set of syntactic codes~$q_d$ that describe the output syntactic
      form at increasing levels granularity. These are combined to
      give a syntactic embedding~$\textbf{z}_{syn}$, which is fed to
      the decoder along with the original semantic content to generate the output sentence~$\textbf{y}$. During training, the encoder is driven by a paraphrase $\textbf{x}_{sem}$ and a syntactic exemplar $\textbf{x}_{syn}$.}
    
    \label{fig:graphical_models}
\end{figure}

In this paper, we propose a generative model for paraphrase generation, that combines the diversity introduced by an explicit syntactic target with the tractability of models trained end-to-end. Shown in \Cref{fig:graphical_models}, the model begins by assuming the existence of some semantic content~$\textbf{z}_{sem}$. Conditioned on this semantic information, the model predicts a syntactic `sketch' in the form of a hierarchical set of discrete codes~$q_{1:D}$, that describe the target syntactic structure with increasing granularity. The sketch is combined into an embedding~$\textbf{z}_{syn}$, and fed along with the original meaning~$\textbf{z}_{sem}$ to a decoder that generates the final output utterance~$\textbf{y}$. Choosing a discrete representation for the sketch means it can be predicted from the meaning as a simple classification task, and the hierarchical nature means that the joint probability over the codes admits an autoregressive factorisation, making prediction more tractable.

The separation between $\textbf{z}_{sem}$ and $\textbf{z}_{syn}$ is induced by a training scheme introduced in earlier work  \cite{hosking-lapata-2021-factorising,huang-chang-2021-generating} and inspired by prior work on separated latent spaces \cite{chen-etal-2019-multi, bao-etal-2019-generating}, whereby the model must reconstruct a target output from one input with the correct meaning, and another input with the correct syntactic form. To learn the discretized sketches, we propose a variant of
Vector-Quantized Variational Autoencoders (VQ-VAE, or VQ) that learns
a \textit{hierarchy of embeddings} within a shared vector space, and represents an input encoding as
a path through this hierarchy. Our approach, which we call
Hierarchical Refinement Quantized Variational Autoencoders or
\textbf{HRQ-VAE}, leads to a decomposition of a dense vector into
embeddings of increasing granularity, representing high-level
information at the top level before gradually refining the encoding
over subsequent levels.



Our contributions are summarized as follows:

\begin{itemize}
    \vspace{-0.2cm}\item We propose a generative model of natural language generation, HRQ-VAE, that induces a syntactic sketch to account for the diversity exhibited by paraphrases. We present a parameterization of our generative model that is a novel method for learning hierarchical discretized embeddings over a single latent encoding space. These embeddings are trained end-to-end and jointly with the encoder/decoder.
    \vspace{-0.2cm}\item We use HRQ-VAE to induce hierarchical
    sketches for paraphrase generation, demonstrating that the known
    factorization over codes makes them easier to predict at
    test time, and leads to higher quality paraphrases.
\end{itemize}

\section{Latent Syntactic Sketches}

\subsection{Motivation}

Let $\textbf{y}$ be a sentence, represented as a sequence of
tokens. We assume that $\textbf{y}$ contains semantic content, that
can be represented by a latent variable~$\textbf{z}_{sem}$. Types of
semantic content might include the description of an image, or a
question intent. However, the mapping from semantics to surface form
is not unique: in general, there is more than one way to express the
semantic content. Sentences with the same underlying
meaning~$\textbf{z}_{sem}$ but different surface form~$\textbf{y}$ are
\textit{paraphrases}. Standard approaches to paraphrasing
(e.g.,~\citealt{bowman-etal-2016-generating}) map directly from
$\textbf{z}_{sem}$ to $\textbf{y}$, and do not account for this
diversity of syntactic structure.

Following recent work on syntax-guided paraphrasing
\cite{chen-etal-2019-controllable,hosking-lapata-2021-factorising},
and inspired by evidence that humans plan out utterances at a higher
level than individual words \cite{martin2010planning}, we introduce an
intermediary \textit{sketching} step, depicted in
\Cref{fig:gm_decoder}. We assume that the output sentence~$\textbf{y}$
is generated as a function both of the meaning~$\textbf{z}_{sem}$
\textit{and} of a syntactic encoding~$\textbf{z}_{syn}$ that describes
the structure of the output. Moreover, since natural language displays
hierarchical organization in a wide range of ways, including at a
syntactic level (constituents may contain other consituents), we also
assume that the syntactic encoding $\textbf{z}_{syn}$ can be
decomposed into a hierarchical set of discrete latent
variables~$q_{1:D}$, and that these $q_d$ are conditioned on the
meaning~$\textbf{z}_{sem}$. This contrasts with
popular model architectures such as VAE \cite{bowman-ppvae} which use a
\emph{flat} internal representation in a dense Euclidean vector space.

Intuitively, our generative model corresponds to a process where a person thinks of a message they wish to convey; then, they decide roughly how to say it, and incrementally refine this decision; finally, they combine the meaning with the syntactic sketch to `spell out' the sequence of words making up the sentence. 





\subsection{Factorization and Objective}

The graphical model in \Cref{fig:gm_decoder} factorizes as
\begin{multline}
    p(\textbf{y}, \textbf{z}_{sem}) 
    = \sum_{q_{1:D},\textbf{z}_{syn}}p(\textbf{y}| \textbf{z}_{sem}, \textbf{z}_{syn}) \\
    \times p(\textbf{z}_{syn} | q_{1:D}) \\ 
\hspace*{-.2cm}\times\hspace*{-.3ex}p(\textbf{z}_{sem})
\hspace*{-.2ex}\times\hspace*{-.2ex} p(q_1 | \textbf{z}_{sem}) \hspace*{-.3ex}\prod
\limits_{d=2}^D \hspace*{-.4ex}\times p(q_d | q_{< d}, \textbf{z}_{sem})
    .
\end{multline}

Although  $q_{1:D}$ are conditionally dependent on
$\textbf{z}_{sem}$, we assume that
$\textbf{z}_{sem}$ may be determined from $\textbf{y}$ without needing
to explicitly calculate~$q_{1:D}$ or $\textbf{z}_{syn}$. We also assume that the mapping
from discrete codes~$q_{1:D}$ to $\textbf{z}_{syn}$ is a deterministic
function~$f_{q \rightarrow \textbf{z}}(\cdot)$. The posterior
therefore factorises as
\begin{multline}
    \phi(\textbf{z}_{sem}, \textbf{z}_{syn} | \textbf{y}) =
     \phi(\textbf{z}_{sem} | \textbf{y}) \times \phi(\textbf{z}_{syn} | \textbf{y})  \\
     \times \phi(q_1 | \textbf{z}_{syn}) \times \prod \limits_{d=2}^D \phi(q_d | q_{< d}, \textbf{z}_{syn}).
\end{multline}


\begin{figure*}[t!]
    \centering
    %
     \includegraphics[width=0.98\textwidth]{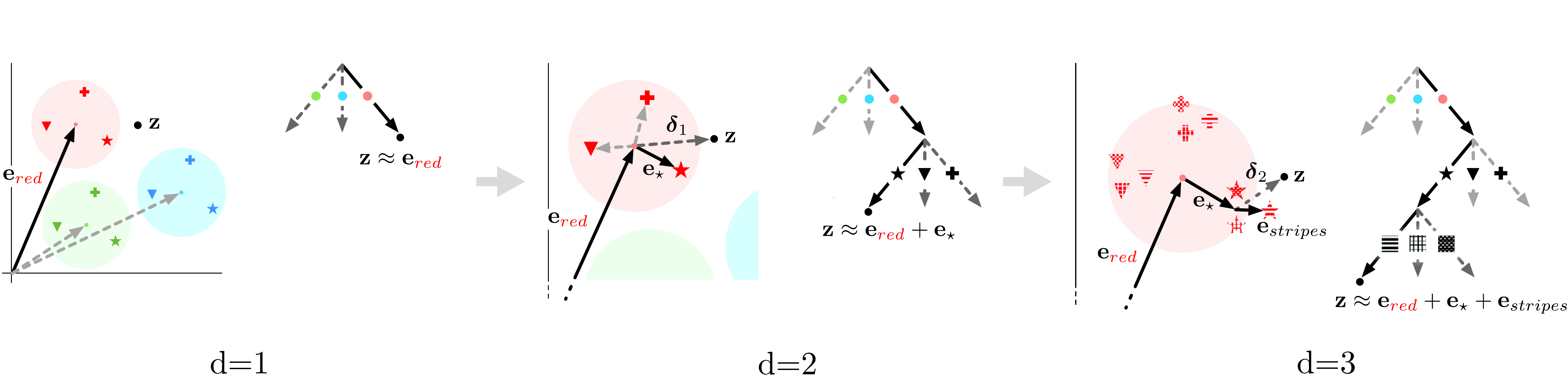}  
    \caption{An illustration of how HRQ-VAE maps an input encoding vector $\textbf{z}$ to a decomposition of hierarchical discretized encodings. HRQ-VAE compares the input to a jointly learned codebook of embeddings that become increasingly granular at lower depths of hierarchy. In this simplified example, with a depth of 3 and a codebook size of 3, the nearest top-level (colours) embedding to $\textbf{z}$ is $\textbf{e}_{\textcolor{red}{red}}$; then, the residual error~$\boldsymbol{\delta}_1 = \textbf{z} - \textbf{e}_{\textcolor{red}{red}}$ is compared to the 2\textsuperscript{nd} level of embeddings (shapes), with the nearest being $\textbf{e}_{\star}$. Finally, the residual error~$\boldsymbol{\delta}_2$ is compared to the 3\textsuperscript{rd} level codebook (patterns), where the closest is $\textbf{e}_{stripes}$. The quantized encoding of $\textbf{z}$ is then  $\textbf{z} \approx \textbf{e}_{\textcolor{red}{red}} + \textbf{e}_{\star} + \textbf{e}_{stripes}$.}
    \vspace{-0.4cm}
    \label{fig:illustration}
\end{figure*}

The separation between $\textbf{z}_{sem}$ and
$q_{1:D}$, such that they represent the meaning and form of the input
respectively, is induced by the
training scheme. During training, the model is trained to reconstruct
a target $\textbf{y}$ using $\textbf{z}_{sem}$ derived from an input with the
correct meaning (a paraphrase) $\textbf{x}_{sem}$, and $q_{1:D}$ from
another input with the correct form (a syntactic exemplar)
$\textbf{x}_{syn}$. \citet{hosking-lapata-2021-factorising} showed that the model therefore learns to encode primarily semantic information about the input in $\textbf{z}_{sem}$, and primarily syntactic information in $q_{1:D}$. Exemplars are retrieved from the training data
following to the process described in 
\citet{hosking-lapata-2021-factorising}, with examples in \Cref{app:exemplars}. The setup is shown in
\Cref{fig:gm_encoder}; in summary, during training we set
$\phi(\textbf{z}_{sem} | \textbf{y}) = \phi(\textbf{z}_{sem} |
\textbf{x}_{sem})$ and $\phi(q_d | \textbf{y}, q_{< d}) = \phi(q_d |
\textbf{x}_{syn}, q_{< d})$. The final objective is given by
\begin{multline} \label{eq:finalobjective}
    \text{ELBO} = \mathbb{E}_{\phi}\big [-\log p(\textbf{y} | \textbf{z}_{sem}, q_{1:D})) \\
     - \log p(q_1 | \textbf{z}_{sem})  - \sum\limits_{d=2}^D \log p(q_d | q_{< d}, \textbf{z}_{sem})  \big ] \\
      + KL\big [\phi(\textbf{z}_{sem} | \textbf{x}_{sem}) || p(\textbf{z}_{sem})\big ],
\end{multline}
where $q_d \sim \phi(q_d|\textbf{x}_{syn})$ and $\textbf{z}_{sem} \sim \phi(\textbf{z}_{sem}|\textbf{x}_{sem} )$.

\section{Neural Parameterisation}

We assume a Gaussian distribution for $\textbf{z}_{sem}$, with prior~$p(\textbf{z}_{sem}) \sim \mathcal{N}(\textbf{0},
\textbf{1})$. The encoders~$\phi(\textbf{z}_{sem} | \textbf{x}_{sem})$ and $\phi(\textbf{z}_{syn} | \textbf{x}_{syn})$ are Transformers \cite{Vaswani2017}, and we use an autoregressive Transformer decoder for $p(\textbf{y} | \textbf{z}_{sem}, \textbf{z}_{syn})$. The mapping~$f_{q \rightarrow \textbf{z}}(\cdot)$ from $q_{1:D}$ to $\textbf{z}_{syn}$ and the posterior network~$\phi(q_d | q_{< d}, \textbf{z}_{syn} )$ are more complex, and form a significant part of our contribution.

Our choice of parameterization is learned end-to-end, and ensures that the sketches learned are hierarchical both in the shared embedding space and in the information they represent. 

\subsection{Hierarchical Refinement Quantization}





Let $\textbf{z}_{syn} \in \mathcal{R}^\mathbb{D}$ be the output of the encoder network $\phi(\textbf{z}_{syn} | \textbf{y})$, that we wish to decompose as a sequence of discrete hierarchical codes. Recall that $q_d \in [1, K]$ are discrete latent variables corresponding to the codes at different levels in the hierarchy, $d \in [1, D]$.
Each level uses a distinct codebook, $\textbf{C}_d \in \mathbb{R}^{K \times \mathbb{D}}$, which maps each discrete code to a continuous embedding $\textbf{C}_d(q_d) \in \mathbb{R}^{\mathbb{D}}$.

 The distribution over codes at each level is a softmax distribution, with the scores~$s_d$ given by the distance from each of the codebook embeddings to the residual error between the input and the cumulative embedding from all previous levels, 
\begin{align}
\hspace{-0.25cm} s_d(q) = - \left( \left[\textbf{x} - \sum\limits_{d'=1}^{d-1} \textbf{C}_{d'}(q_{d'}) \right ] - \textbf{C}_d(q)  \right ) ^2\hspace{-0.25cm} .
\end{align}
Illustrated in \Cref{fig:illustration}, these embeddings therefore represent iterative refinements on the quantization of the input. The posterior network~$\phi(q_d | q_{< d}, \textbf{z}_{syn})$ iteratively \textit{decomposes} an encoding vector into a path through a hierarchy of clusters whose centroids are the codebook embeddings.


Given a sequence of discrete codes $q_{1:D}$, we deterministically construct its continuous representation with the composition function $f_{q \rightarrow \textbf{z}}(\cdot)$,
\begin{align}
\textbf{z}_{syn} = f_{q \rightarrow \textbf{z}}(q_{1:D}) = \sum\limits_{d=1}^{D} \textbf{C}_d(q_{d}).
\end{align}

HRQ-VAE can be viewed as an extension of VQ-VAE \cite{vqvae}, with two
significant differences: (1) the codes are hierarchically ordered and
the joint distribution $p(q_1, \ldots, q_D)$ admits an autoregressive
factorization; and (2) the HRQ-VAE composition function is a sum,
compared to concatenation in VQ or a complex neural network in VQ-VAE 2 \cite{vqvae2}. Under HRQ, latent codes describe
a path through the learned hierarchy within a shared encoding space. The form of the posterior~$\phi(q_d
| q_{< d}, \textbf{z}_{syn} )$ and the composition function~$f_{q
  \rightarrow \textbf{z}}(\cdot)$ do not rely on any particular
properties of the paraphrasing task; the technique could be applied to
any encoding space.





\paragraph{Initialisation Decay} Smaller perturbations in encoding space should result in more fine grained changes in the information they encode. Therefore, we encourage \textit{ordering} between the levels of hierarchy (such that lower levels encode finer grained information) by initialising the codebook with a decaying scale, such that later embeddings have a smaller norm than those higher in the hierarchy. Specifically, the norm of the embeddings at level $d$ is weighted by a factor~$(\alpha_{init}) ^ {d-1}$.


\paragraph{Depth Dropout} To encourage the hierarchy within the encoding space to correspond to hierarchical properties of the output, we introduce \textit{depth dropout}, whereby the hierarchy is truncated at each level during training with some probability $p_{depth}$. The output of the quantizer is then given by 
\begin{align}
\textbf{z}_{syn} = \sum \limits_{d=1}^D \left ( \textbf{C}_d(q_d) \prod\limits_{d'=1}^d \gamma_{d'} \right ) ,
\end{align}
where $\gamma_h \sim \text{Bernoulli}(1 - p_{depth})$. This means that the model is sometimes trained to reconstruct the output based only on a \textit{partial} encoding of the input, and should learn to cluster similar outputs together at each level in the hierarchy.

\subsection{Sketch Prediction Network}

During training the decoder is driven using sketches sampled from the encoder, but at test time exemplars are unavailable and we must predict a distribution over syntactic sketches $p(q_{1:D} |\textbf{z}_{sem})$. Modelling the sketches as hierarchical ensures that this distribution admits an autoregressive factorization.


We use a simple recurrent network to infer valid codes at each level of hierarchy, using the semantics of the input sentence and the cumulative embedding of the predicted path so far as input, such that
$q_d$ is sampled from  $p(q_d |\textbf{z}_{sem}, q_{< d}) = \texttt{Softmax}(\texttt{MLP}_d(\textbf{z}_{sem}, \textbf{z}_{<d}))$, where $\textbf{z}_{<d} = \sum\limits_{d'=1}^{d-1} \textbf{C}_{d'}(q_{d'})$. This MLP is trained jointly with the encoder/decoder model, using the outputs of the posterior network $\phi(q_d |
\textbf{x}_{syn}, q_{< d})$ as targets.
To generate paraphrases as test time, we sample from the sketch prediction model $p(q_d |\textbf{z}_{sem}, q_{< d})$ using beam search and condition generation on these predicted sketches.



\begin{figure*}[ht!]
    \centering
    \includegraphics[width=1\textwidth]{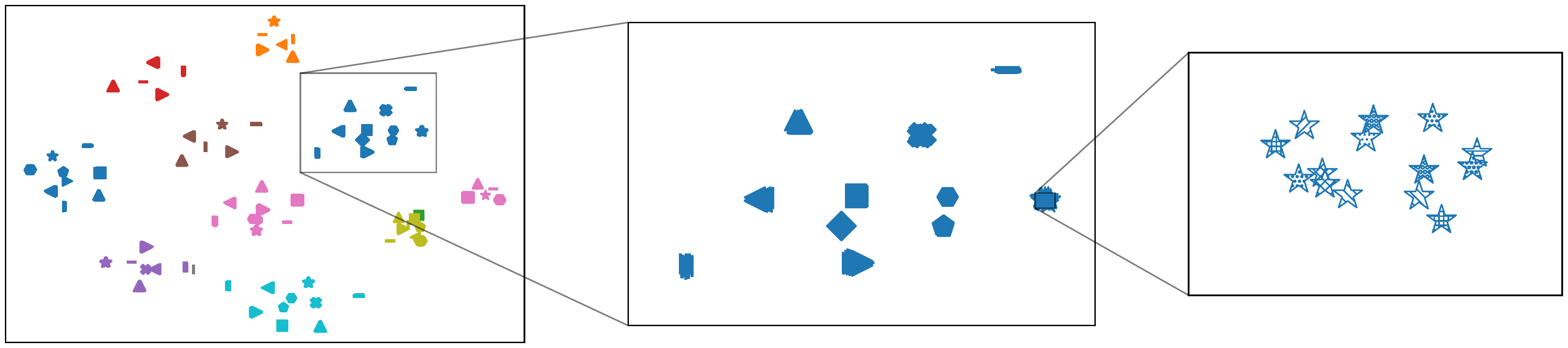}
    \caption{t-SNE visualisation of the syntactic encodings $\textbf{z}_{syn}$ for 10k examples from Paralex: colours indicate top-level codes $q_1$, shapes indicate the second level, and patterns are used to label the third level. Deeper levels in the hierarchy represent finer grained information in encoding space.}
    \vspace{-0.4cm}
    \label{fig:tsne}
\end{figure*}

\subsection{Training Setup}
\label{sec:training}

We use the Gumbel reparameterisation trick
\cite{jang2016categorical,maddison2017concrete,sonderby2017continuous}
for the discrete codes and the standard Gaussian reparameterisation
for the semantic representation. To encourage the model to use the full codebook, we decayed the Gumbel temperature~$\tau$, according to the schedule given in \Cref{app:hyperparams}. We approximate the expectation in
\Cref{eq:finalobjective} by sampling from the training set and
updating via backpropagation \cite{kingma2013autoencoding}. The full model was trained jointly by optimizing the ELBO in \Cref{eq:finalobjective}.



\section{Experimental Setup}

\paragraph{Datasets}

A paraphrase is `an alternative surface form in the
same language expressing the same semantic content as the original
form' \cite{madnanidorr}, but it is not always clear what counts as the `same semantic content'. Our approach requires access to reference paraphrases; we evaluate on three English paraphrasing datasets which have clear grounding for the meaning of each sentence: Paralex
\cite{fader-etal-2013-paraphrase}, a dataset of question paraphrase
clusters scraped from WikiAnswers; Quora Question Pairs
(QQP)\footnote{\mbox{\url{https://www.kaggle.com/c/quora-question-pairs}}}
sourced from the community question answering forum Quora; and MSCOCO 2017 \cite{mscoco}, a set of images that have been captioned by multiple annotators. For the question datasets, each paraphrase is grounded to the (hypothetical) \textit{answer} they share. We use the splits released by \citet{hosking-lapata-2021-factorising}. For MSCOCO, each caption is grounded by the \textit{image} that it describes. We evaluate on the public validation set, randomly selecting one caption for each image to use as input and using the remaining four as references.

\paragraph{Model Configuration}

Hyperparameters were tuned on the Paralex development set, and reused for the other evaluations. We set the depth of the hierarchy $D = 3$, and the codebook size $K = 16$. The Transformer encoder and decoder consist of 5 layers each, and we use the vocabulary and token embeddings from BERT-Base \cite{devlin_bert:_2018}. We use an initialisation decay factor of $\alpha_{init} = 0.5$, and a depth dropout probability $p_{depth} = 0.3$. A full set of hyperparameters is given in \Cref{app:hyperparams}, and our code is available at {\url{ https://github.com/tomhosking/hrq-vae}}.

\paragraph{Comparison Systems}


As baselines, we consider three popular architectures: a vanilla autoencoder (AE) that learns a single dense vector representation of an input sentence; a Gaussian Variational AutoEncoder \citep[VAE,][]{bowman-ppvae}, which learns a distribution over dense vectors; and a Vector-Quantized Variational AutoEncoder \citep[VQ-VAE,][]{vqvae}, that represents the full input sentence as a set of discrete codes. All three models are trained to generate a sentence from one of its paraphrases in the training data, and are not trained with an autoencoder objective. We implement a simple \mbox{tf-idf} baseline
\cite{tfidf}, retrieving the question from the training set with the
highest cosine similarity to the input. Finally, we include a basic copy baseline as a lower bound, that simply uses the input sentences as the output.

We also compare to a range of recent paraphrasing systems. Latent
bag-of-words \citep[BoW,][]{latentbow} uses an encoder-decoder model
with a discrete bag-of-words as the latent encoding. SOW/REAP
\cite{goyal_neural_2020} uses a two stage approach, deriving a set of
feasible syntactic rearrangements that is used to guide a second
encoder-decoder model. BTmPG \cite{lin-wan-2021-pushing} uses
multi-round generation to improve diversity and a reverse paraphrasing
model to preserve semantic fidelity. We use the results after 10
rounds of paraphrasing. Separator
\cite{hosking-lapata-2021-factorising} uses separated,
non-hierarchical encoding spaces for the meaning and form of an input,
and an additional inference model to predict the target syntactic form
at test time. All comparison systems were trained and evaluated on our
splits of the datasets.

As an upper bound, we select a sentence from the evaluation set to use as an \textit{oracle} syntactic exemplar, conditioning generation on a sketch that is known to represent a valid surface form.

\begin{table*}[ht!]
    \centering
\small
    \begin{tabular}{l|rrr|rrr|rrr}
     & \multicolumn{3}{c|}{\textbf{Paralex}} & \multicolumn{3}{c|}{\textbf{QQP}} & \multicolumn{3}{c}{\textbf{MSCOCO}} \\
    \textbf{System} & {BLEU} $\uparrow$ & {Self-B} $\downarrow$  & {\textbf{iBLEU}} $\uparrow$  & {BLEU} $\uparrow$  & {Self-B} $\downarrow$ & {\textbf{iBLEU}} $\uparrow$ & {BLEU} $\uparrow$  & {Self-B} $\downarrow$ & {\textbf{iBLEU}} $\uparrow$ \\
\hline \hline
    Copy  & 37.10 & 100.00 & 9.68 & 34.52 & 100.00 & 7.61 & 19.85 & 100.00 & -4.12 \\ 
    \mbox{tf-idf}  & 25.08 & 25.25 & 15.01 & 24.05 & 62.49 & 6.75 & 18.26 & 38.37 & 6.93 \\ 
    AE  & 40.10 & 75.71 & 16.94 & 28.99 & 60.11 & 11.17 & 27.90 &  38.71 & 14.58 \\ 
    VAE  & 38.91 & 53.28 & 20.47 & 27.23 & 51.09 & 11.57 & 27.44 & 24.40 & 16.99 \\

    VQ-VAE  & 40.26 & 65.71 & 19.07 & 16.31 & 21.13 & 8.83 & 25.62 & 22.41 & 16.01 \\ 
    \hline
    SOW/REAP & 33.09 & 37.07 & 19.06 & 21.27 & 38.01 & 9.41 & 12.51 & 6.47 & 8.71 \\ 
    LBoW & 34.96 & 35.86 & 20.80 & 23.51 & 42.08 & 10.39 & 21.65 & 16.46 & 14.02 \\ 
    BTmPG & 28.40 & 35.99 & 15.52 & 19.83 & 35.11 & 8.84 & 19.76 & 13.04 & 13.20 \\

    Separator & 36.36 & 35.37 & 22.01 & 23.68 & 24.20 & 14.10 & 20.59 & 12.76 & 13.92  \\
    HRQ-VAE & 39.49 & 33.30 & \textbf{24.93} & 33.11 & 40.35 & \textbf{18.42} &  27.90 & 16.58 & \textbf{19.04}  \\
    \hline 
    {Oracle} & 50.58 & 28.09 & 34.85 & 50.47 & 36.84 & 33.01 & 35.80 & 12.85 & 26.07 \\

    \hline \hline
    \end{tabular}
    \caption{Top-1 paraphrase generation results, without access to oracle
      sketches. HRQ-VAE achieves the highest iBLEU scores, indicating the best tradeoff between quality and diversity. Paired bootstrap resampling \cite{koehn-2004-statistical} indicates that HRQ-VAE significantly improves on all other systems (p$ < 0.05$).}  
    \label{tab:ibleu}
\end{table*}

\section{Results}

Our experiments were designed to test two primary hypotheses: (1) Does HRQ-VAE learn \textit{hierarchical} decompositions of an encoding space? and (2) Does our choice of generative model enable us to generate\textit{ high quality} and \textit{diverse} paraphrases? 


\subsection{Probing the Hierarchy}
\label{sec:probing-hierarchy}

\Cref{fig:tsne} shows a t-SNE \cite{tsne} plot of the syntactic encodings $\textbf{z}_{syn}$ for 10,000 examples from Paralex. The encodings are labelled by their quantization, so that colours indicate top-level codes $q_1$, shapes denote $q_2$, and patterns $q_3$. The first plot shows clear high level structure, with increasingly fine levels of substructure visible as we zoom into each cluster. This confirms that the discrete codes are ordered, with lower levels in the hierarchy encoding more fine grained information. 

To confirm that intermediate levels of hierarchy represent valid
points in the encoding space, we generate paraphrases using oracle
sketches, but truncate the sketches at different depths. Masking one level (i.e.,~using only $q_1,q_2$) reduces performance by~$2.5$ iBLEU points,
and two levels by~$5.5$. (iBLEU is an automatic metric for assessing
paraphrase quality; see Section~\ref{sec:paraphr-gener}). Although
encodings using the full depth are the most informative,
\textit{partial} encodings
still lead to good quality output, with a gradual degradation. This implies both that each level
in the hierarchy contains useful information, and that the cluster
centroids at each level are representative of the individual members
of those clusters.




\subsection{Paraphrase Generation}
\label{sec:paraphr-gener}


\paragraph{Metrics}

Our primary metric is iBLEU \cite{ibleu},
\begin{align}
\begin{split}
    \textrm{iBLEU} = \alpha \textrm{BLEU}(outputs, references) \\- (1-\alpha) \textrm{BLEU}(outputs, inputs),
\end{split}
\end{align}
that measures the fidelity of generated outputs to reference paraphrases as well as the level of diversity introduced. We use the corpus-level variant. Following the recommendations of \citet{ibleu}, we set $\alpha = 0.8$, with a sensitivity analysis shown in \Cref{app:hyperparams}.  We also report $\textrm{BLEU}(outputs, references)$ as well as
\mbox{Self-{BLEU}}$(outputs, inputs)$. The latter allows us to examine the extent to which models generate paraphrases that differ from the original input.

To evaluate the diversity between multiple candidates generated by the \textit{same system}, we report pairwise-BLEU \cite{cao-wan-2020-divgan},
\begin{align*}
\begin{split}
    \textrm{P-BLEU} = \mathbb{E}_{i \neq j} [ \textrm{BLEU}(outputs_i, outputs_j) ].
\end{split}
\end{align*}
This measures the average similarity between the different candidates, with a lower score indicating more diverse hypotheses.

\paragraph{Automatic Evaluation}

Shown in \Cref{tab:ibleu}, the results of the automatic evaluation highlight the importance of measuring both paraphrase quality and similarity to the input: the Copy baseline is able to achieve high BLEU scores despite simply duplicating the input. The VAE baseline is competitive but tends to have a hi\begin{table}[t]
    \small
    \centering
    \begin{tabular}{@{}r@{~}|@{~}p{6cm}@{}}
    \hhline{==} 
    \textit{Paralex} & Where is the birthplace of woman pro golfer Dottie Pepper? \\
    \hline
    {VAE} & Where is the birthplace of Pepper pro golfer Dottie? \\
    {BTmpG} & What is the birthplace of women pro golfer? \\
    {SOW/REAP} &  What is the birthplace for golfer? \\
    {Latent BoW} & Where did the golfer golfer originate? \\
    {Separator} &  Where is the birthplace of Dottie? \\
    {HRQ-VAE} &  Where is Dottie Pepper from? \\
    \hhline{==} 	 
    \textit{QQP} & What are the best ways to defrost lobster tails? \\
    \hline
    {VAE} &   What are the best ways to defrost lobster tails? \\
    {BTmpG} &  How can I defrost my tails?? \\
    {SOW/REAP} & What is defrost? \\
    {Latent BoW} &  How do you something a something lobster? \\
    {Separator} &  What are some of the best ways to defrost chicken? \\
    {HRQ-VAE} & How do you thaw frozen lobster tails? \\
    \hhline{==}
    
    \textit{MSCOCO} & Set of toy animals sitting in front of a red wooden wagon. \\
    \hline
    {VAE} &  Two stuffed animals sitting in front of a toy train. \\
    {BTmpG} &   A herd of sheep grazing in a field of grass. \\
    {SOW/REAP} & A close up of a close up of a street \\
    {Latent BoW} & A toy wagon with a toy horse and a toy wagon. \\
    {Separator} &  A toy model of a toy horse and buggy. \\
    {HRQ-VAE} & A group of stuffed animals sitting next to a wooden cart. \\
    \hhline{==}
    \end{tabular}
    \caption{Examples of generated paraphrases. HRQ-VAE is able to preserve the original meaning, while introducing significant syntactic variation.}
    
    \label{tab:examples} 
    
\end{table}

gh Self-BLEU score, indicating that the semantic preservation comes at the cost of low syntactic diversity. HRQ-VAE achieves both higher BLEU scores and higher iBLEU scores than the comparison systems, indicating that it is able to generate higher quality paraphrases without compromising on syntactic diversity.

The examples in \Cref{tab:examples} demonstrate
that HRQ is able to introduce significant syntactic variation while
preserving the original meaning of the input. However, there is still
a gap between generation using predicted sketches and `oracle' sketches
(i.e.,~when the target syntactic form is known in advance), indicating ample scope for improvement.

\begin{table}[t]
\small
    \centering
    \begin{tabular}{@{~}c@{~}c@{~}c@{~}|@{~}p{6cm}@{}}
    \hline \hline
    $q_1$ & $q_2$ & $q_3$ & \textbf{Output} \\
    \hline
    \multicolumn{3}{@{~}c@{~}|@{~}}{\textit{Input}} & Two types of fats in body ? \\
    \hline
\multirow{2}{*}{0} & 3 & 6  &  What types of fats are in a body?  \\
\cline{2-4}
 & 13 & 7  &  What types of fats are there in body?  \\
\hline
\multirow{2}{*}{2} & 1 & 2  &  How many types of fats are there in the body?  \\
\cline{2-4}
 & 3 & 7  &  How many types of fats are there in a body?  \\
\hline
\multirow{4}{*}{5} & 3 & 6  &  What are the different types of fats in a body?  \\
\cline{2-4}
 & 5 & 7  &  What are the different types of fats in body?  \\
\cline{2-4}
 & \multirow{2}{*}{8} & 7  &  Types of fats are different from body fat?  \\
 &  & 14  &  Two types of fats in body?  \\
\hline
\multirow{6}{*}{13} & \multirow{2}{*}{0} & 2  &  What are the different types of fats in the body?  \\

 &  & 6  &  What are the different types of fats in a body?  \\
\cline{2-4}
 & 3 & 7  &  What are two types of fats in a body?  \\

\cline{2-4}
 & \multirow{3}{*}{5} & 7  &  What are the different types of fats in body?  \\

 &  & 8  &  What are the different types of fats?  \\

 &  & 14  &  What are the different types of fats in the body?  \\


\hline \hline
    \end{tabular}
    \caption{Examples of model output, for a range of different sketches. The left hand side shows the sketch (i.e.,~the values of the codes $q_{1:D}$), with the corresponding model output on the right. $q_1$ primarily specifies the wh- word (e.g., outputs with $q_1=13$ are all `what' questions), while $q_2,q_3$ correspond to more fine grained details, e.g.,~the outputs with $q_3=6$ all use the article `a' when referring to `body'.}
    \label{tab:workedex}
\end{table}

\paragraph{Worked Example}

Since the sketches $q_{1:D}$ are latent variables, interpretation is difficult. However, a detailed inspection of example output reveals some structure.

\Cref{tab:workedex} shows the model output for a single semantic input drawn from Paralex, across a range of different syntactic sketches. It shows that $q_1$ is primarily responsible for encoding the question type, with $q_1=13$ leading to `what' questions and $q_1=2$ `how' questions. $q_2$ and $q_3$ encode more fine grained details; for example, all outputs shown with $q_3=6$ use the indefinite article `a'.

We also examine how using increasingly granular sketches refines the syntactic template of the output. \Cref{tab:maskedexample} shows the model output for a single semantic input, using varying granularities of sketch extracted from the exemplar. When no sketch is specified, the model defaults to a canonical phrasing of the question. When only $q_1$ is specified, the output becomes a `how many' question, and when a full sketch is included, the output closely resembles the exemplar.

\begin{table}[t]
\small
    \centering
    \begin{tabular}{@{~}c@{~}|@{~}p{5.9cm}@{}}
    
    \hline \hline
    \textit{Input} & Two types of fat in body? \\
    \textit{Exemplar} & How many states are in the USA? \\
    \hline
    No sketch & What are the different types of fats in the body?  \\
    $q_1$ & How many types of fats are there in the body? \\
    $q_1,q_2$ & How many fats does the body have? \\
    $q_1,q_2,q_3$ & How many fat are in the body? \\
    \hline
    \hline
    \end{tabular}
    \caption{Model output for varying sketch granularities. When no sketch is used, the model defaults to the most common phrasing of the question. As more detail is included, the output converges towards the exemplar.}
    \label{tab:maskedexample}
\end{table}

\paragraph{Generating Multiple Paraphrases} We evaluated the ability of our system to generate multiple diverse paraphrases for a single input, and compared to the other comparison systems capable of producing more than one output. For both HRQ-VAE and Separator, we used beam search to sample from the sketch prediction network as in the top-1 case, and condition generation on the top-3 hypotheses predicted. For BTmPG, we used the paraphrases generated after 3, 6 and 10 rounds. For the VAE, we conditioned generation on 3 different samples from the encoding space. The results in \Cref{tab:topk} show that HRQ-VAE is able to generate multiple high quality paraphrases for a single input, with lower similarity between the candidates than other systems.

\begin{table}[t!]
    \centering
\small
    \begin{tabular}{@{}l@{\hspace{.1em}}|@{\hspace*{.1em}}r@{\hspace{.2em}}@{}r@{\hspace{.1em}}|@{\hspace*{.1em}}r@{\hspace{.2em}}@{}r@{\hspace{.1em}}|@{\hspace*{.1em}}r@{\hspace{.2em}}@{}r@{\hspace{.1em}}}
     & \multicolumn{2}{c@{}|@{\hspace*{.1em}}}{\textbf{Paralex}} & \multicolumn{2}{c@{}|@{\hspace*{.1em}}}{\textbf{QQP}} & \multicolumn{2}{c}{\textbf{MSCOCO}} \\
     
    \textbf{Model}  & \rot{iBLEU $\uparrow$} & \rot{P-BLEU $\downarrow$} & \rot{iBLEU $\uparrow$} &  \rot{P-BLEU $\downarrow$} & \rot{iBLEU $\uparrow$} & \rot{P-BLEU $\downarrow$} \\
\hline \hline
    VAE & 20.49 & 67.62 & 11.52 & 64.71  & 17.22 & 55.66   \\
    BTmPG & 15.50 & 89.20 & 9.13 & 82.02 & 13.20 & 80.38   \\
    Separator & 21.67 & 62.98 & 13.63 & \textbf{52.87} & 13.77 & 57.79 \\
      {HRQ-VAE} & \textbf{22.75} & \textbf{40.48} & \textbf{17.49} & 57.29 & \textbf{18.39} & \textbf{41.29} \\
    \hline \hline
    \end{tabular}
    \caption{Top-3 generation results. P-BLEU indicates the similarity between the different candidates, while iBLEU scores reported are the mean across the 3 candidates. HRQ-VAE is able to generate multiple high quality paraphrases with more diversity between them than comparison systems.}  
    \label{tab:topk}
\end{table}

\begin{figure}[t]
    \centering
    \includegraphics[width=0.49\textwidth]{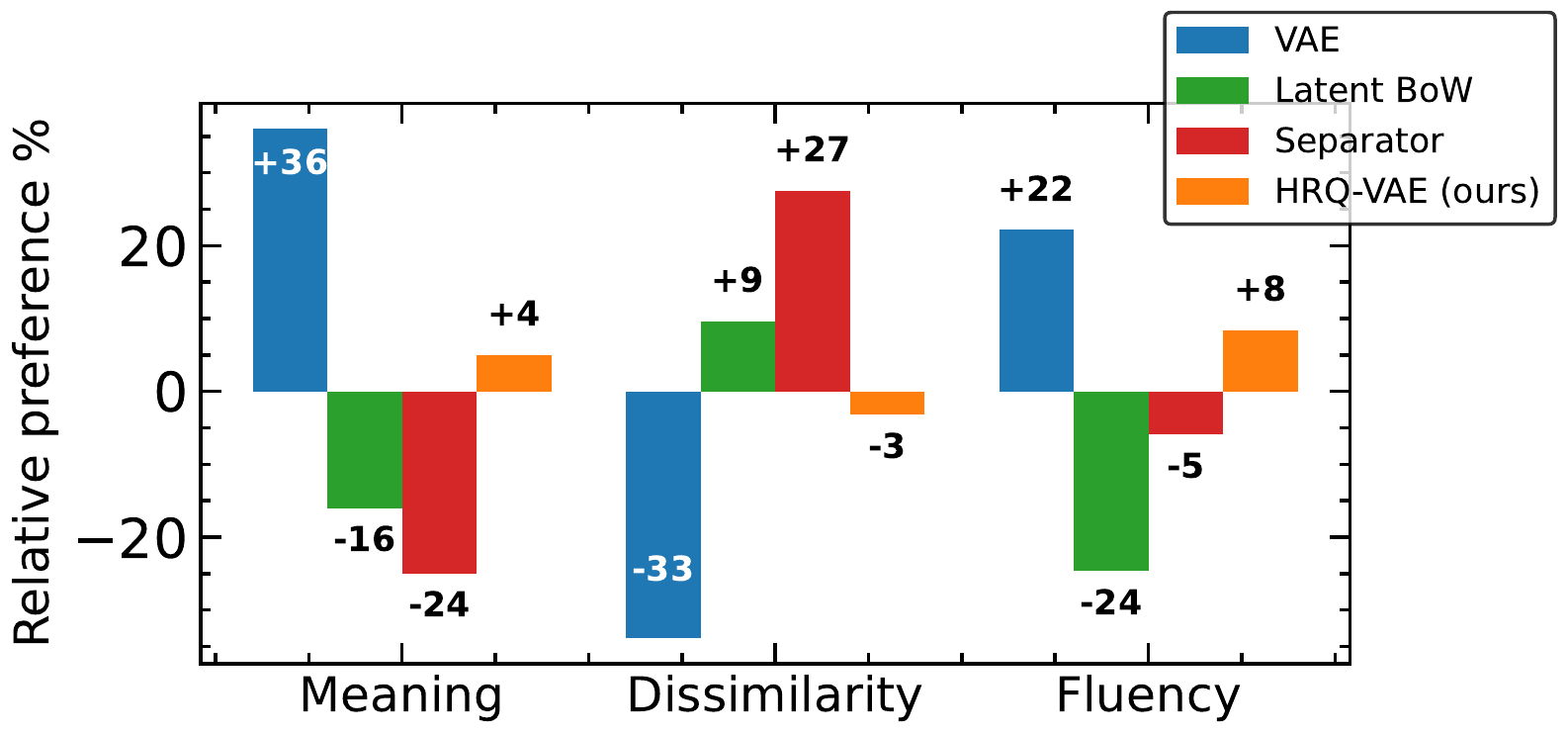}
    \caption{Results of our human evaluation. Although the VAE
      baseline is the best at preserving sentence meaning, it is the
      worst at introducing variation to the output. HRQ-VAE
      offers the best balance between dissimilarity and meaning
      preservation, and is more fluent than both Separator and Latent BoW. } 
    \label{fig:humaneval}
\end{figure}

\subsection{Human Evaluation}

In addition to automatic evaluation we elicited judgements from
crowdworkers on Amazon Mechanical Turk. They were shown
a sentence and two paraphrases, each generated by a different system, and asked to select which one was preferred along three
dimensions: the \textit{dissimilarity} of the paraphrase compared to
the original sentence; how well the paraphrase reflected the
\textit{meaning} of the original; and the \textit{fluency} of the
paraphrase (see \Cref{app:humeval}). We evaluated a total of 300
sentences sampled equally from each of the three evaluation datasets, and collected 3
ratings for each sample. We assigned each system a score of $+1$ when
it was selected, $-1$ when the other system was selected, and took the
mean over all samples. Negative scores indicate that a system was
selected less often than an alternative. We chose the four best
performing models for our evaluation:
HRQ-VAE, Separator, Latent BoW, and VAE.

\Cref{fig:humaneval} shows that although the VAE baseline is the best
at preserving question meaning, it is also the worst at introducing
variation to the output. HRQ-VAE better preserves the original
question intent compared to the other systems while introducing more diversity than the VAE, as well as generating much more fluent
output.

\subsection{Ablations}

To confirm that the hierarchical model allows for more expressive sketches, we performed two ablations. We compared to the full model using oracle sketches, so that code prediction performance was not a factor. We set the depth $D=1$ and $K=48$, giving equivalent total capacity to the full model ($D=3,K=16$) but without hierarchy. We also removed the initialisation scaling at lower depths, instead initialising all codebooks with the same scale. \Cref{tab:ablations} shows that a non-hierarchical model with the same capacity is much less expressive.

We also performed two ablations against the model using predicted sketches; we removed depth dropout, so that the model is always trained on a full encoding. We confirm that learning the codebooks jointly with the encoder/decoder leads to a stronger model, by first training a model with a continuous Gaussian bottleneck (instead of the HRQ-VAE); then, we recursively apply $k$-means clustering \cite{kmeans}, with the clustering at each level taking place over the residual error from all levels so far, analogous to HRQ-VAE. The results of these ablations shown in \Cref{tab:ablations} indicate that our approach leads to improvements over all datasets.

\begin{table}[t!]
    \centering
\small
    \begin{tabular}{l@{~}|r@{~}|r@{~}|r@{}}
     \textbf{Variant} & {\textbf{Paralex}} & {\textbf{QQP}} & {\textbf{MSCOCO}} \\
\hline \hline
    HRQ-VAE (oracle) & 34.85 & 33.01 & 26.07  \\
    \hline
    No initialisation scaling & $-$3.06 & $-$2.48 & $-$3.02 \\
    No hierarchy & $-$8.84 & $-$12.72 & $-$3.10 \\
    \hline \hline
    HRQ-VAE & 24.93 & 18.42 & 19.04  \\
    \hline 
    No head dropout & $-$0.62 & $-$0.74 & $-$0.81 \\
    
    Post-hoc k-means & $-$3.30 & $-$5.35 & $-$2.83 \\
    \hline \hline
    \end{tabular}
    \caption{Changes in iBLEU score for a range of ablations from our
      full model. All components lead to an improvement in paraphrase
      quality across datasets.} 
    \label{tab:ablations}
\end{table}


\section{Related Work}

\paragraph{Hierarchical VAEs}

VQ-VAEs were initially proposed in computer vision \cite{vqvae}, and were later extended to be `hierarchical' \cite{vqvae2}. However, in vision the term refers to a `stacked' version architecture, where the output of one variational layer is passed through a CNN and then another variational layer that can be continuous \cite{vahdat2020NVAE} or quantized \cite{hqa,Livin2019TowardsHD,willetts2021relaxedresponsibility}. Unlike these approaches, we induce a \emph{single} latent space that has hierarchical properties. 

Other work has looked at using the properties of hyperbolic geometry to encourage autoencoders to learn hierarchical representations. \citet{mathieu2019poincare} showed that a model endowed with a Poincaré ball geometry was able to recover hierarchical structure in datasets, and \citet{suris2021hyperfuture} used this property to deal with uncertainty in predicting events in video clips. However, their work was limited to continuous encoding spaces, and the hierarchy discovered was known to exist a priori. 

\paragraph{Syntax-controlled Paraphrase Generation}



Prior work on paraphrasing has used retrieval techniques \cite{barzilay-mckeown-2001-extracting}, Residual LSTMs \cite{prakash-etal-2016-neural}, VAEs \cite{bowman-etal-2016-generating}, VQ-VAEs
\cite{roy-grangier-2019-unsupervised} and pivot languages \cite{mallinson-etal-2017-paraphrasing}. Syntax-controlled paraphrase generation has seen significant recent interest, as a means to explicitly generate diverse surface forms with the same meaning. However, most previous work has required knowledge of the correct or valid surface forms to be generated \cite{iyyer-etal-2018-adversarial,chen-etal-2019-controllable,sgcp2020,meng2021conrpg}. It is generally assumed that the input can be rewritten without addressing the problem of predicting which template should be used, which is necessary if the method is to be useful. \citet{hosking-lapata-2021-factorising} proposed learning a simplified representation of the surface form using VQ, that could then be predicted at test time. However, the discrete codes learned by their approach are not independent and do not admit a known factorization, leading to a mismatch between training and inference. 



\section{Conclusion}

We present a generative model of paraphrasing, that uses a hierarchy of discrete latent variables as a rough syntactic sketch. We introduce HRQ-VAE, a method for mapping these hierarchical sketches to a continuous encoding space, and demonstrate that it can indeed learn a hierarchy, with lower levels representing more fine-grained information. We apply HRQ-VAE to the task of paraphrase generation, representing the syntactic form of sentences as paths through a learned hierarchy, that can be predicted during testing. Extensive experiments across multiple datasets and a human evaluation show that our method leads to high quality paraphrases. The generative model we introduce has potential application for any natural language generation task; $\textbf{z}_{sem}$ could be sourced from a sentence in a different language, from a different modality (e.g.,~images or tabular data) or from a task-specific model (e.g.,~summarization or machine translation). 
Furthermore, HRQ-VAE makes no assumptions about the type of space being represented, and could in principle be applied to a semantic space, learning a hierarchy over words or concepts.

\section*{Acknowledgements}

We thank our anonymous reviewers for their feedback. This work was supported in part by the
UKRI Centre for Doctoral Training in Natural Language Processing,
funded by the UKRI (grant EP/S022481/1) and the University of
Edinburgh. Lapata acknowledges the support of the European Research
Council (award number 681760, ``Translating Multiple Modalities into
Text'').

\bibliography{anthology,separator,main}
\bibliographystyle{acl_natbib}

\appendix


\section{Hyperparameters}
\label{app:hyperparams}

The hyperparameters given in \Cref{tab:hyperparams} were selected by manual tuning, based on a combination
of: (a)~validation iBLEU scores with depth masking, (b)~validation BLEU scores
using oracle sketches, and (c)~validation iBLEU scores using
predicted syntactic codes.

\begin{table}[ht]
    \centering
\small
    \begin{tabular}{l|p{2cm}}
    \textbf{Encoder/decoder} & \\
    Embedding dimension $D$ & 768 \\
    Encoder layers & 5 \\
    Decoder layers & 5 \\
    Feedforward dimension & 2048 \\
    Transformer heads & 8 \\
    Semantic/syntactic dim & 192/594 \\
    Depth $D$& 3 \\
    Codebook size $K$& 16 \\
    Optimizer & Adam \cite{adam} \\
    Learning rate & 0.01 \\
    Batch size & 64 \\
    Token dropout & 0.2 \cite{tokendropout} \\
    Decoder & Beam search \\
    Beam width & 4 \\
    \hline
    \textbf{Code predictor} & \\
    Num. hidden layers & 2 \\
    Hidden layer size & 3072 \\
    
    \end{tabular}
    \caption{Hyperparameter values used for our experiments.}
    \label{tab:hyperparams}
\end{table}

The Gumbel temperature $\tau$ is decayed during training as a function of the step $t$, according to the following equation:
\begin{equation}
  \tau(t) = \max(2 - \frac{2}{1 + e^{t/10000}}, 0.5) .
\end{equation}
Intuitively, this smoothly decays $\tau$ from an initial value of 2, with a half-life of 10k steps, to a minimum value of 0.5.

\begin{figure*}[t!]
    \centering
    \includegraphics[width=0.95\textwidth]{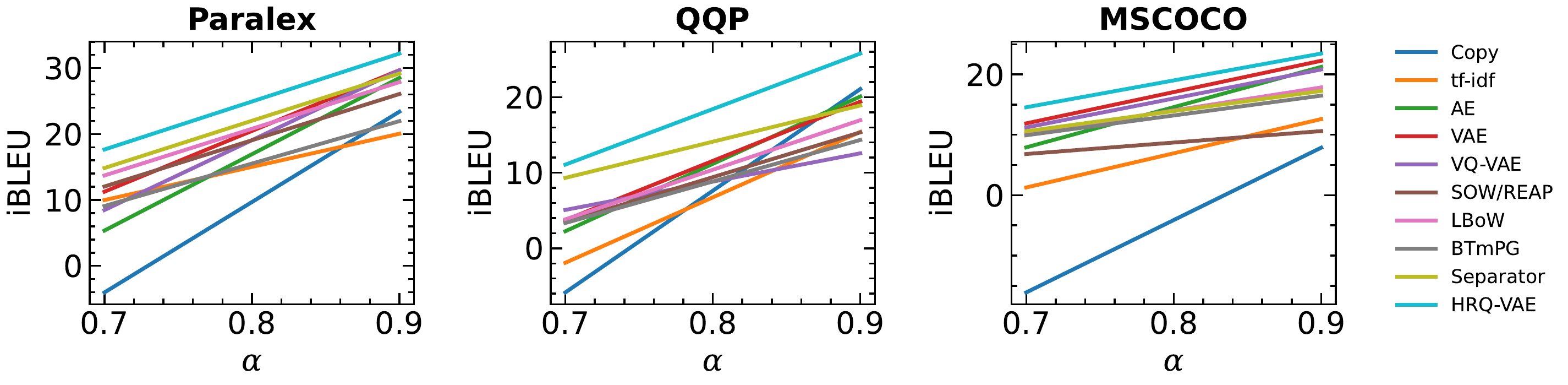}
    \vspace{-0.2cm}
    \caption{iBLEU scores for all comparison systems, for a range of values of $\alpha$.}
    \label{fig:ibleusens}
\end{figure*}

We use $\alpha=0.8$ when calculating iBLEU, but as shown in \Cref{fig:ibleusens} our conclusions are not sensitive to this value, and our model outperforms all comparison systems on all datasets for $0.7 \leq \alpha \leq 0.9$.

Models were trained on a single GPU, with training taking between one and three days depending on the dataset. We use SacreBLEU \cite{post-2018-call} to calculate BLEU scores.

\section{Human Evaluation}
\label{app:humeval}

Annotators were recruited from the UK and USA via Amazon Mechanical Turk, and were compensated for their time above a living wage in those countries. A full Participant Information Sheet was provided, and the study was approved by an internal ethics committee. Annotators were asked to rate the outputs according to the following criteria:

\begin{itemize}
    \item Which system output is the most fluent and grammatical?
    \item To what extent is the meaning expressed in the original sentence preserved in the rewritten version, with no additional information added?
    \item Does the rewritten version use different words or phrasing to the original? You should choose the system that uses the most different words or word order.
\end{itemize}

\section{Exemplar Retrieval Process}
\label{app:exemplars}

\begin{table}[t!]
    \small
    \centering
    \begin{tabular}{@{}c@{~}|@{~}c@{}}
    
    \hhline{==}
        \textit{Input} & How heavy is a moose? \\
        \textit{Chunker output} & How [heavy]\textsubscript{ADVP} is a [moose]\textsubscript{NP} ? \\
        \textit{Template} & How ADVP is a NP ? \\
        \textit{Exemplar} & How much is a surgeon's income? \\
    \hhline{==}
        \textit{Input} & What country do parrots live in \\
        \textit{Chunker output} & What [country]\textsubscript{NP} do  [parrots]\textsubscript{NP} [live]\textsubscript{VP} in ? \\
        \textit{Template} & What NP do NP VP in ? \\
        \textit{Exemplar} & What religion do Portuguese believe in? \\
     \hhline{==}
    \end{tabular}
    \caption{Examples of the exemplar retrieval process for training. The input is tagged by a chunker, ignoring stopwords. An exemplar with the same template is then retrieved from a different paraphrase cluster. Table reproduced with permission from \citet{hosking-lapata-2021-factorising}.}
    \label{tab:templateexample}
\end{table}

Our approach requires exemplars during training
to induce the separation between latent spaces. We follow the approach introduced by \citet{hosking-lapata-2021-factorising}. During training, we retrieve exemplars $\textbf{x}_{syn}$ from the
training data following a process which first identifies the
underlying syntax of $\textbf{Y}$, and finds a question with the same
syntactic structure but a different, arbitrary meaning. We use a
shallow approximation of syntax, to ensure the availability of
equivalent exemplars in the training data. An example of the exemplar
retrieval process is shown in \Cref{tab:templateexample}; we first
apply a chunker \citep[FlairNLP, ][]{akbik-etal-2018-contextual} to
$\textbf{Y}$, then extract the chunk label for each tagged span,
ignoring stopwords. This gives us the \textit{template} that
$\textbf{Y}$ follows. We then select a question at random from the
training data with the same template to give $\textbf{x}_{syn}$. If no
other questions in the dataset use this template, we create an exemplar by replacing each chunk with a random sample of the same type.


\section{Analysis of Code Properties}

\begin{figure}[t!]
    \centering
    \begin{subfigure}{0.44\textwidth}
        \raggedleft
        \includegraphics[width=0.97\textwidth]{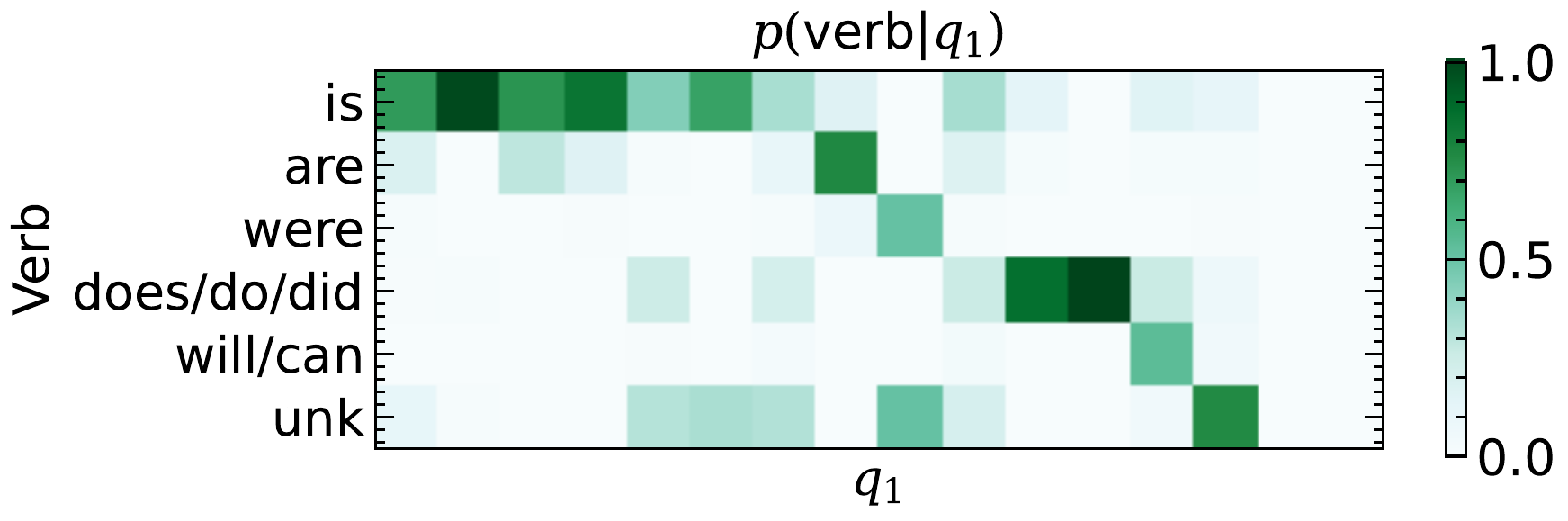}
        \caption{Distribution of verbs for each code within level 1.}
        \label{fig:verbs}
    \end{subfigure} %
    \begin{subfigure}{0.46\textwidth}
        \raggedleft
        \includegraphics[width=0.86\textwidth]{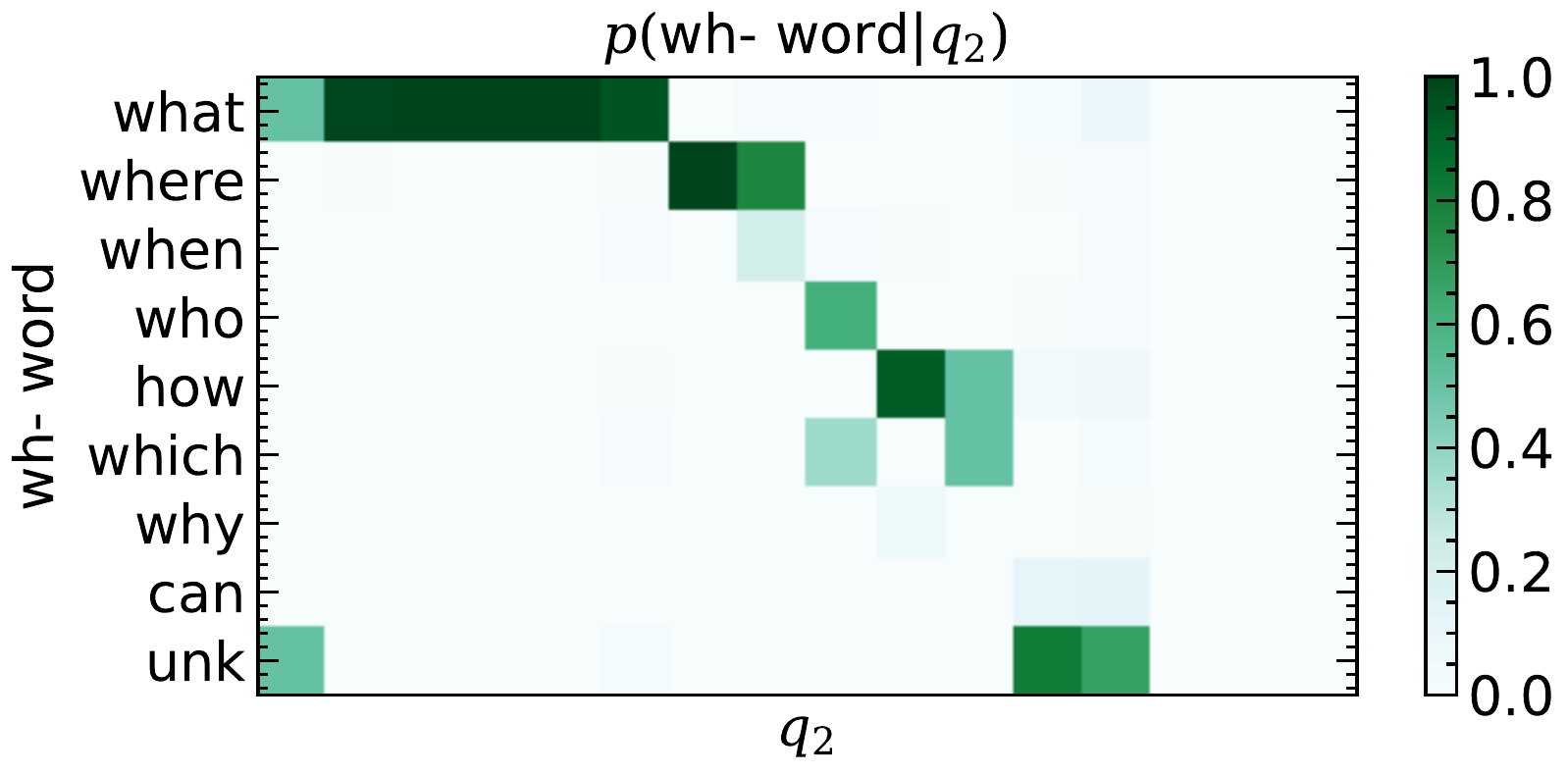}
        \caption{Distribution of wh- words for each code within level 2.}
        \label{fig:whwords}
    \end{subfigure}

    \caption{Plots showing the conditional distributions of two
      different sentence features, auxiliary verb and question type,
      for different values of the latent codes~$q_d$. Each column
      represents the distribution over the feature for a specific
      code. The plots show that level 1 is a strong predictor of verb
      tense, and level 2 predicts question type, giving some insight
      into what syntactic features each level has learned to
      encode. We have reordered the columns of the plot to improve
      readability.}
    \label{fig:heads}
\end{figure}

We define two
features of sentences: (1) the presence of common auxiliary verbs that
roughly indicate the tense of the sentence (present, future, etc.);
and (2) the presence of different question or `wh-'
words\footnote{This analysis was performed for Paralex, which
  comprises entirely of questions.}. We calculate the distributions of
these features for each code $q_d$ at different levels, with the
results shown in \Cref{fig:heads}. Each column represents the
distribution over the feature for a specific code. \Cref{fig:verbs}
shows clear evidence that the sentences are (at least partly)
clustered at the top level based on the verb used, while
\Cref{fig:whwords} shows that level 2 encodes the question type.

\end{document}